\documentclass{article}

\PassOptionsToPackage{numbers,compress,sort&compress}{natbib}

\usepackage[main, final]{neurips/neurips_2026}

\usepackage[utf8]{inputenc} 
\usepackage[T1]{fontenc}    
\usepackage{hyperref}       
\usepackage{url}            
\usepackage{booktabs}       
\usepackage{amsfonts}       
\usepackage{amsmath}
\usepackage{amssymb}
\usepackage{mathtools}
\usepackage{amsthm}
\usepackage{nicefrac}       
\usepackage{microtype}      
\usepackage{xcolor}         
\usepackage{graphicx}
\usepackage{subcaption}
\usepackage{placeins}
\usepackage{algorithm}
\usepackage{algorithmic}
\usepackage[capitalize,noabbrev]{cleveref}
\usepackage[textsize=tiny]{todonotes}

\theoremstyle{plain}

\theoremstyle{definition}

\theoremstyle{remark}

\crefname{appendix}{Appendix}{Appendices}
\Crefname{appendix}{Appendix}{Appendices}

\title{WELP : Weighted Ensemble in the LooP for committor learning}

\author{%
  Jacky Chen$^{1,2}$ \qquad Rishal Aggarwal$^{1,2}$ \qquad David Ryan Koes$^{2}$ \\
  \\
  $^{1}$Carnegie Mellon - University of Pittsburgh PhD Program in Computational Biology\\
  $^{2}$Department of Computational \& Systems Biology\\
  University of Pittsburgh\\
  Pittsburgh, PA 15260
}

\begin{document}
\maketitle
\begin{abstract}
Understanding how complex molecular systems transition between metastable states is a central challenge in computational chemistry and biophysics. A key tool for characterizing these transitions is the committor function: the probability that a trajectory initiated from a given configuration reaches one state before another. Because the committor encodes both the progress and the mechanism of a transition, it is an ideal collective variable (CV). Learning it, however, requires sampling the rare transitions it describes. We address this circular dependency with a ``WE-in-the-loop'' framework that couples committor learning with weighted ensemble (WE), an enhanced sampling method that uses a CV to drive transition dynamics. Our framework alternates between the two: WE simulation generates transition data for training a neural network committor with a self-consistent loss, and the learned committor then serves as an improved CV for subsequent WE simulation. Across a two-channel 2D system, alanine dipeptide, and chignolin, our framework consistently outperforms greedy baselines, highlighting the importance of principled exploration. We also curate a public benchmark of empirically determined committor values for alanine dipeptide. Together, the framework and benchmark provide a new method for studying rare transitions in high-dimensional molecular systems and a standard for evaluating committor learning methods.
\end{abstract}

\section{Introduction}
Many important processes in chemistry and biophysics are governed by rare transitions between metastable states. Understanding the mechanisms of such transitions is a standing challenge. Sampling transitions with molecular dynamics (MD) simulations is often computationally intractable as the system spends the vast majority of its time in metastable basins, with transition events occurring on timescales longer than what unbiased simulation can access \cite{shaw_atomic-level_2010}. This has long motivated the development of enhanced sampling and transition path sampling methods that aim to make observing these transitions tractable \cite{bolhuis_transition_2002,bogetti_suite_2019}.

One key tool to characterize these transition paths is the committor function. For any given configuration, the forward committor function defines the probability that a trajectory initiated from that configuration will reach the target state before the source state. The committor is the ideal reaction coordinate to sample along as it is the most informative collective variable, encoding complete kinetic information about the transition pathways \cite{e_transition-path_2010}. Furthermore, knowing the committor allows one to calculate a variety of different properties of the transition path ensemble, as well as accelerate enhanced sampling \cite{menzl_s-shooting_2016}.  However, computing the committor requires sampling in the transition regions, which is exactly the hardest region to access, creating a chicken-and-egg problem.

It is possible to parametrize the committor function using a neural network \cite{giuseppe_chen_following_2026,rotskoff_active_2021}. However, these methods still need data in transition regions, which is hard to obtain. 
\citet{mitchell_committor_2024} aimed to alleviate this issue by alternating between sampling along an imperfect committor to collect data and training the committor on this freshly collected data. At each round, the neural network outputs the committor value for a set of points in configuration space, and the point with the largest predicted committor value is chosen to launch N new trajectories from. The committor is then updated using the results from these N new trajectories and the process is repeated. This approach is greedy / argmax-style: at each iteration, the next 
starting point is the point with the largest predicted committor value under the current estimate. However, in related fields such as reinforcement learning, greedy sampling has been shown to converge to suboptimal solutions \cite{sutton_reinforcement_nodate}. Similarly, poor performance of greedy algorithms has also been observed in traditional transition path sampling \cite{zhang_weighted_2009-1}.

Principled simulation frameworks that provide better exploration-exploitation tradeoffs correct for the biases of greedy sampling. One such algorithm, weighted ensemble (WE), is based on maintaining a set of bins across a reaction coordinate. Trajectories (called walkers) are then simulated in an unbiased fashion and are assigned to bins. As walkers fluctuate between bins, constant splitting and merging of trajectories is done to keep a consistent number of trajectories within each bin~\cite{huber_weighted-ensemble_1996,zhang_weighted_2009}, encouraging uniform sampling along the reaction coordinate. Weighted ensemble is widely used and has several notable successful applications including determining rates of protein folding \cite{adhikari_computational_2019}, the mechanisms of membrane transporters \cite{adelman_simulations_2011}, and the RBD opening pathways of the SARS-CoV-2 Spike protein \cite{sztain_glycan_2021}.

Weighted ensemble (WE) sampling has been successful at simulating transition paths in large molecular dynamics systems, but its efficiency depends on an informative reaction coordinate. Since the committor is the ideal reaction coordinate, we ask two complementary questions: \emph{can WE be used to learn the committor, and can a learned committor in turn accelerate WE?}

We show that the answer to both is yes, and that the two goals reinforce each other: WE generates the rare transition data needed to train a committor model, and the learned committor provides an increasingly informative coordinate for guiding WE. Building on this observation, we make the following contributions:

\begin{itemize}
    \item \textbf{Method.} We develop a WE-in-the-loop algorithm that alternates between WE simulation and neural network committor training, using the learned committor to define adaptive bins for WE resampling.
    \item \textbf{Validation.} We show that our method outperforms a greedy baseline on three systems of increasing complexity: a 2D multi-channel potential; alanine dipeptide (ADP), a 22-atom peptide that serves as the standard benchmark for rare-event sampling methods; and chignolin, a fast-folding protein in explicit solvent.
    \item \textbf{Dataset.} We curate and release a benchmark dataset of empirically determined committor values for ADP, constructed through systematic clustering and brute-force trajectory shooting.
\end{itemize}

We believe these contributions, together with the open-source release of our code and datasets, will help accelerate research in rare-event sampling.

\section{Related Work}

Committor learning methods differ in both their objectives and how they acquire training data. \citet{khoo_solving_2018} use a variational formulation of the committor equation, minimizing the equilibrium-weighted expected squared gradient of the committor subject to its boundary conditions. Because this objective weights samples by their equilibrium probability, it requires data from the rarely visited transition region. \citet{chen_discovering_2023} also take a variational approach, training a Siamese neural network to find the slowest decorrelating collective variable, which approximates the committor. \citet{aristoff_fast_2024} avoid neural networks entirely, representing the committor with a kernel method. \citet{strahan_predicting_2023} instead use a self-consistency objective: since the committor at a configuration equals the expected committor at the end of a short trajectory started from it, the model can be trained on short trajectories alone rather than full transitions. This idea forms the basis of the semigroup method of \citet{mitchell_committor_2024}. Other methods focus on how data is acquired. \citet{rotskoff_active_2021} use active importance sampling to concentrate data in the transition region, while \citet{jung_machine-guided_2023} use transition path sampling (TPS), with an active-learning scheme to select better initial configurations. 

From the weighted ensemble perspective, previous learned-coordinate approaches have pursued different targets than the committor. \citet{wang_augmenting_2024} combine human-based and data-based bins to define better collective variables for WE. \citet{leung_unsupervised_2025} combine a VAE with latent outlier detection to bias additional starting points.  \citet{yang_rare-event_2024} continuously apply clustering to identify novel outliers to initialize trajectories from. None of these methods trains a committor, and they often require expensive procedures such as clustering or human augmented CVs.      

Our work connects these two lines: we train with a self-consistent objective and acquire data with weighted ensemble, using the learned committor to direct sampling.



\section{Methodological Framework}
\begin{figure}
    \centering
    \includegraphics[width=1\linewidth]{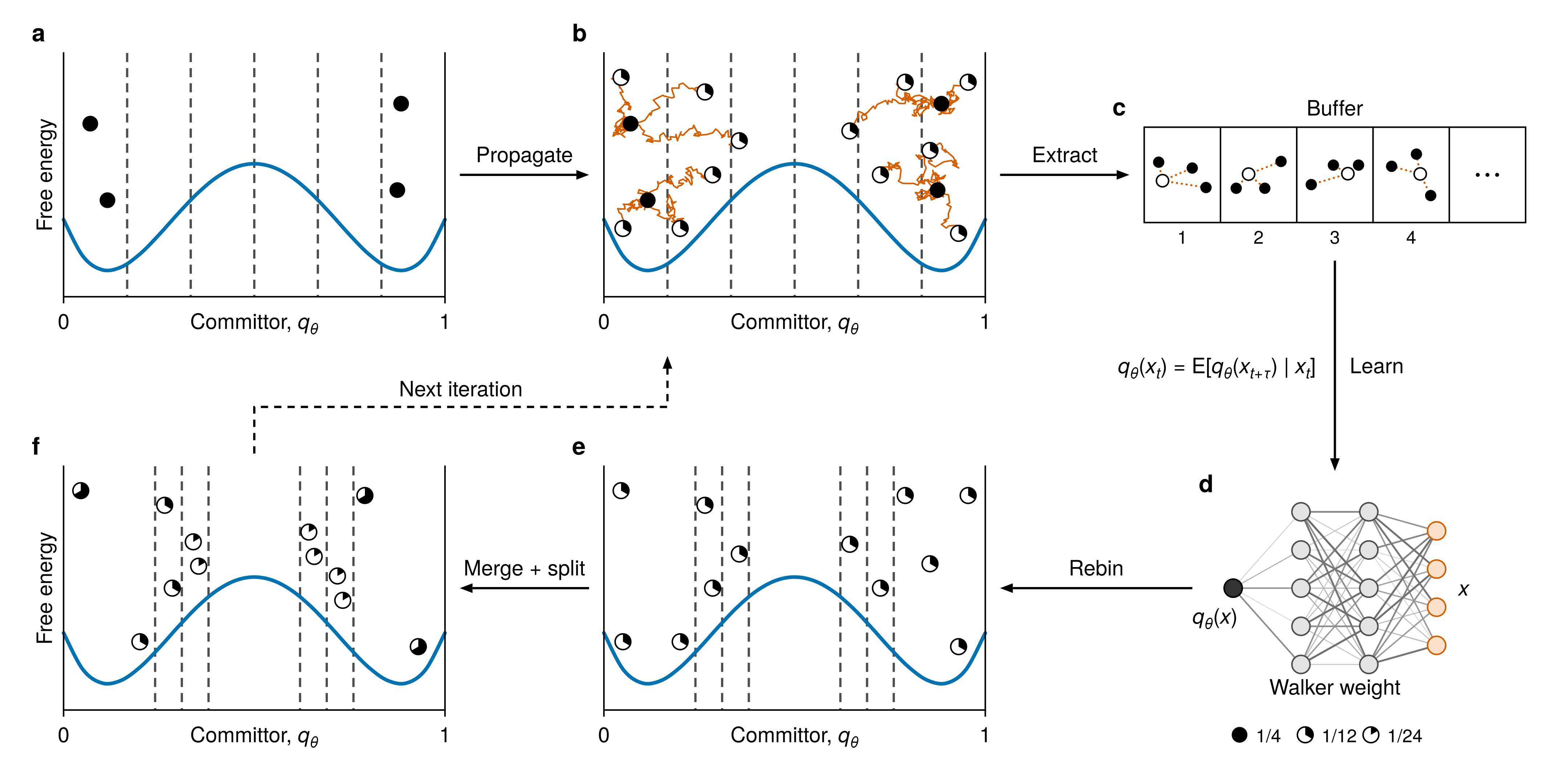}
    \caption{\textbf{Algorithmic overview of WELP}, with a target of two walkers per bin: (a) initialize walkers at boundary states; (b) propagate the dynamics; (c) store parent–child pairs in a buffer; (d) train the neural network using the semigroup loss; (e) adaptively rebin to increase sampling in bottleneck regions; and (f) merge and split walkers to restore two walkers per bin.}
    \label{fig:overview}
\end{figure}
\subsection{Motivation}
A key challenge in committor learning is that the model's training data depends on where we choose to sample. A natural strategy is greedy: concentrate new samples where the current committor estimate is the highest, as in the seeding scheme of \citet{mitchell_committor_2024}. However, this makes the method brittle. Sampling remains concentrated in regions the current estimate already considers important, alternative pathways are never visited, and the model cannot correct its own mistakes. This failure mode is well known in reinforcement learning as the exploration--exploitation tradeoff, where acting greedily on the current value estimate can prevent an agent from discovering better solutions \citep{sutton_reinforcement_nodate}, and it has also been observed in path sampling \citep{zhang_weighted_2009-1}. Our WE-in-the-loop framework avoids this by training on samples from WE, which balances both sides of the tradeoff. Adaptive binning maintains walkers in every bin, ensuring broad coverage of configuration space (exploration), while splitting trajectories in bottleneck regions focuses effort where transitions occur (exploitation). When choosing a loss function to to learn on top of our WE simulation, we hypothesized that the semigroup loss would be uniquely suited for this purpose.

This pairing of WE and the semigroup loss is not arbitrary; the two components reinforce each other. WE generates many short trajectories rather than a single long continuous path, which is exactly the data the semigroup loss is designed to use. In turn, the learned committor makes WE more efficient, since binning along the true reaction coordinate is what determines resampling quality.

\subsection{Background: Langevin Dynamics and Rare Events}

We model molecular systems as stochastic processes in which atoms move under a potential energy $U$ while exchanging energy with a surrounding heat bath at inverse temperature $\beta = 1/k_B T$. This exchange is described by Langevin dynamics, and we consider two regimes depending on whether particle momenta are modeled explicitly.

When friction is large, momenta equilibrate much faster than the positional dynamics of interest and can be adiabatically eliminated. This gives overdamped Langevin dynamics:
\begin{equation} \label{eq:overdamped}
    dX_t = -\nabla U(X_t)\,dt + \sqrt{2\beta^{-1}}\,dW_t,
\end{equation}
where $X_t \in \Omega \subseteq \mathbb{R}^d$ is the configuration, $U\colon \Omega \to \mathbb{R}$ is the potential energy, and $W_t$ is a standard $d$-dimensional Brownian motion. Here the state of the system is fully specified by the position $X_t$. We use this regime for the 2D benchmark, integrated with the Euler--Maruyama scheme, and for the adversarial alanine dipeptide experiments.

For realistic biomolecular simulations, momenta must be retained, leading to underdamped Langevin dynamics:
\begin{equation} \label{eq:underdamped}
\begin{aligned}
    dX_t &= M^{-1}P_t\,dt, \\
    dP_t &= -\nabla U(X_t)\,dt - \gamma M^{-1}P_t\,dt + \sqrt{2\gamma\beta^{-1}}\,dW_t,
\end{aligned}
\end{equation}
where $P_t \in \mathbb{R}^d$ is the momentum, $M$ is the mass matrix, and $\gamma$ is the friction coefficient. The state is now the pair $(X_t, P_t)$. 

In both regimes, metastable states are separated by high energy barriers, so transitions between them are rare on simulation timescales. Obtaining transition paths is therefore a rare-event sampling problem, and characterizing these transitions motivates the committor function introduced below.


\paragraph{Committor Functions.}
\label{sec:committor-def}
Given the dynamics, and the metastable states $ A,B  \subset \Omega$, we can define the hitting times as $\tau_A = \inf\{t > 0 : X_t \in A\}, \quad \tau_B = \inf\{t > 0 : X_t \in B\}$, where $\Omega$ is the space of configurations. In other words, $\tau_{A/B}$ corresponds to the smallest $t$ before the system enters state $A/B$. The forward committor can then be defined as \cite{e_towards_2006}:
\begin{equation}\label{eq:committor_def}
    q^+(x) = \mathbb{P}(\tau_B < \tau_A \mid X_0 = x)
\end{equation}
That is, the committor $q^+(x)$ is the probability that a trajectory started from configuration $x$ reaches state $B$ before state $A$. By definition, $q^+(x) = 0$ for $x \in A$ and $q^+(x) = 1$ for $x \in B$. The committor is the optimal one-dimensional reaction coordinate, as it exactly captures the progress of transitions between the two states \cite{e_towards_2006, e_transition-path_2010}. The backward committor $q^-(x)$ is defined analogously as the probability that a trajectory arriving at $x$ last visited $A$ rather than $B$; for reversible dynamics, $q^- = 1 - q^+$. Since the backward committor is then fully determined by the forward one, we write $q := q^+$ throughout.

\paragraph{The Semigroup Approach.}
\label{sec:semigroup}
The semigroup approach \citep{li_semigroup_2022, mitchell_committor_2024} trains the committor using a self-consistency property of the dynamics, rather than requiring complete transition paths. Let $T = \inf\{t \geq 0 : X_t \in A \cup B\}$ be the first time a trajectory reaches either basin. For any lag time $\tau > 0$, the committor satisfies
\begin{equation}\label{eq:semigroup_identity}
    q(x) = \mathbb{E}\bigl[\, q(X_{\tau \wedge T}) \mid X_0 = x \,\bigr],
\end{equation}
where $\tau \wedge T = \min(\tau, T)$. In words, the probability of reaching $B$ before $A$ from $x$ equals the average committor value a short time later. Trajectories that enter a basin before time $\tau$ are stopped there and contribute $0$ (if they reach $A$) or $1$ (if they reach $B$).

We turn this identity into a training target for a neural network $q_\theta$. Define the boundary-corrected model
\begin{equation}
    \bar q_\theta(z) =
    \begin{cases}
        0 & z \in A, \\
        1 & z \in B, \\
        q_\theta(z) & \text{otherwise},
    \end{cases}
\end{equation}
and the target
\begin{equation}\label{eq:semigroup_target}
    y_\tau(x) = \mathbb{E}\bigl[\, \bar q_\theta(X_{\tau \wedge T}) \mid X_0 = x \,\bigr]
    \approx \frac{1}{K} \sum_{k=1}^{K} \bar q_\theta\bigl(X^{(k)}_{\tau \wedge T^{(k)}}\bigr),
\end{equation}
where $X^{(k)}$ are $K$ independent short trajectories started from $x$. The true committor is a fixed point of this map: if $q_\theta = q$, then $y_\tau = q_\theta$.

We train $q_\theta$ to match this target with the loss
\begin{equation}\label{eq:semigroup_loss}
\begin{aligned}
    \mathcal{L}_{\mathrm{sg}}(\theta) = \mathbb{E}_{x \sim \mu}\Bigl[ & \bigl(\log q_\theta(x) - \log \mathrm{sg}[y_\tau(x)]\bigr)^2 \\
    & + \bigl(\log(1 - q_\theta(x)) - \log(1 - \mathrm{sg}[y_\tau(x)])\bigr)^2 \Bigr],
\end{aligned}
\end{equation}
where $\mu$ is the distribution of starting configurations and $\mathrm{sg}[\cdot]$ denotes the stop-gradient operator. Two design choices are worth explaining. First, the stop-gradient treats $y_\tau(x)$ as a fixed regression label at each step, so gradients flow only through $q_\theta(x)$. This is analogous to target networks in temporal-difference learning, and it prevents the model from lowering the loss by moving the target toward its own prediction. Second, the log-space residual measures relative rather than absolute error. Committor values near the basins can be very close to $0$ or $1$, and a plain squared error would largely ignore errors there. The two symmetric terms penalize relative errors in both $q_\theta$ and $1 - q_\theta$, so neither basin is favored. In practice, we clip $y_\tau$ to $[\epsilon, 1-\epsilon]$ to keep the logarithms finite. A derivation is given in \cref{app:loss-semigroup}.

\paragraph{Weighted Ensemble and Adaptive Binning.}
\label{sec:we}

Weighted ensemble (WE) \citep{huber_weighted-ensemble_1996} evolves an ensemble of $N$ trajectories, or walkers, each carrying a statistical weight $w_i$, with $\sum_{i=1}^{N} w_i = 1$. Configuration space is partitioned into bins along a chosen progress coordinate, and each bin is assigned a target number of walkers. After each iteration, walkers are resampled to restore this target. In bins with too few walkers, walkers are split into $n$ copies, each receiving an equal share of the parent's weight. In bins with too many, walkers are merged: one walker is kept, chosen with probability proportional to its weight, and it receives the combined weight of the merged group. Both operations conserve the total weight, so the ensemble remains statistically unbiased while computational effort is redistributed toward sparsely sampled regions. We use the WESTPA implementation \citep{russo_westpa_2022}.

Because WE remains unbiased for any choice of bins, the bins need not be placed by hand and can instead be adapted during the simulation. The Minimal Adaptive Binning (MAB) scheme \citep{torrillo_minimal_2021} repositions the bins at each iteration based on the current walker distribution, and adaptive binning of this kind has been shown empirically to speed up convergence. MAB assigns a dedicated bin to the bottleneck walker, defined as the walker $i$ that maximizes
\begin{equation}\label{eq:mab_bottleneck}
    \log w_i - \log \sum_{j \succ i} w_j,
\end{equation}
where $j \succ i$ denotes walkers farther along the progress coordinate. This criterion selects walkers that carry substantial weight relative to the total weight ahead of them, which typically lie on the uphill slopes of free-energy barriers. Because each bin maintains its target walker count, a bottleneck walker in its own bin is repeatedly split rather than merged, concentrating computation on the configurations that drive forward progress.

\paragraph{Steady-state simulations.} WE can be run either in equilibrium or in a non-equilibrium steady state. We use the steady-state setup, which targets the $A \to B$ or $B \to A$ transition directly. Walkers are initialized in the start state and whenever a walker reaches the end state, it is removed and replaced by a new walker started from the start state with the same weight. This recycling conserves the total weight and, over many iterations, drives the ensemble to a steady state with a constant probability flux from the start state to the end state. 

\subsection{WE in the Loop}
\label{sec:committor-in-loop}
In this work, we develop a closed-loop approach that alternates between training a neural network committor $q_\theta$ and using it to adaptively bin weighted ensemble simulations. We run two steady-state WE simulations; $A \to B$ and $B \to A$, so that training data covers both sides of the transition region. Outside the basins, both simulations are binned with MAB using only the predicted committor $q_\theta$ as the progress coordinate.

Each WE iteration also generates the data required by the semigroup loss. Within each bin, all walkers are merged into a single parent, which is then split into $K$ children with freshly sampled velocities. Each child is propagated for a lag time $\tau$, or until it enters a basin, so the children form exactly the short rollouts used to estimate the target $y_\tau$ in \cref{eq:semigroup_target}. Each parent and its children's endpoints are stored in a replay buffer $\mathcal{B}$. After every $T$ WE iterations, we train $q_\theta$ for $E$ epochs on $\mathcal{B}$ ($E = 120$ for the molecular systems and $E = 80$ for the 2D system), computing targets with a target network $q_{\theta^-}$ that is refreshed after each training round. The updated committor then defines the bins for the next round. The full procedure is given in \cref{alg:cil}, and system-specific details (optimizer, replay buffer, target network, and architecture) are provided in \cref{app:training-implementation,app:2d-we,app:adp-we}.

\begin{algorithm}[!htbp]
\caption{WE-in-the-loop committor learning.}
\label{alg:cil}
\begin{algorithmic}[1]
\REQUIRE Committor network $q_\theta$; WE simulations $\mathcal{S}_{A\to B}, \mathcal{S}_{B\to A}$; iterations per round $T$; training epochs $E$; children per bin $K$; lag time $\tau$; learning rate $\eta$.
\STATE $\mathcal{B} \leftarrow \emptyset$; \quad $\theta^- \leftarrow \theta$
\REPEAT
  \FOR{$t = 1$ to $T$}
    \FOR{$\mathcal{S} \in \{\mathcal{S}_{A\to B}, \mathcal{S}_{B\to A}\}$}
      \STATE $\mathcal{G} \leftarrow \mathrm{MAB}(\mathcal{S}, q_\theta)$ \hfill $\triangleright$ bins along $q_\theta$
      \FOR{each bin $g \in \mathcal{G}$ with walkers $\{(x_i, w_i)\}_{i \in g}$}
        \STATE $w_g \leftarrow \sum_{i \in g} w_i$; \quad draw $x_g = x_i$ with probability $w_i / w_g$ \hfill $\triangleright$ merge
        \FOR{$k = 1$ to $K$}
          \STATE Sample velocities $v^{(k)} \sim \mathcal{N}(0, \beta^{-1} M^{-1})$ \hfill $\triangleright$ split
          \STATE $x_g^{(k)} \leftarrow \mathrm{Propagate}(x_g, v^{(k)}, \tau \wedge T^{(k)})$; \quad weight $w_g / K$
        \ENDFOR
        \STATE $\mathcal{B} \leftarrow \mathcal{B} \cup \{(x_g, \{x_g^{(k)}\}_{k=1}^{K})\}$
      \ENDFOR
      \STATE Recycle children that reached the target basin
    \ENDFOR
  \ENDFOR
  \FOR{$e = 1$ to $E$}
    \STATE Sample mini-batch $\{(x_b, \{x_b^{(k)}\}_{k=1}^{K})\}_{b} \sim \mathcal{B}$
    \STATE $y_b \leftarrow \frac{1}{K}\sum_{k=1}^{K} \bar q_{\theta^-}(x_b^{(k)})$ \hfill $\triangleright$ \cref{eq:semigroup_target}
    \STATE $\theta \leftarrow \theta - \eta \nabla_\theta \mathcal{L}_{\mathrm{sg}}(\theta; \{x_b, y_b\})$ \hfill $\triangleright$ \cref{eq:semigroup_loss}
  \ENDFOR
  \STATE $\theta^- \leftarrow \theta$
\UNTIL{converged}
\end{algorithmic}
\end{algorithm}

\paragraph{Balancing exploration and exploitation.} Using the learned committor as the WE progress coordinate balances exploration and exploitation. For exploitation, MAB places bins along $q_\theta$, so bottleneck walkers near the current estimate of the transition region are repeatedly split, concentrating computation where the committor indicates progress. For exploration, WE keeps walkers in every occupied bin across the full range of $q_\theta$, and merging is stochastic and weight-proportional rather than an $\arg\max$ selection. Regions that the current estimate undervalues, including alternative pathways, therefore continue to be sampled. These samples supply training data exactly where $q_\theta$ is wrong, so the model can correct itself rather than collapsing onto a single pathway. Because split and merge conserve walker weights, this broader coverage comes without sacrificing statistical correctness. To our knowledge, this is the first work to pair weighted ensemble sampling with a semigroup committor loss.

\paragraph{Greedy baseline.} We compare against the committor-greedy method of \citet{mitchell_committor_2024}, using the same semigroup loss, architecture, and learning rate as in their paper and code. We refer to this method as \textbf{Q-greedy} throughout.

\section{Experiments}
We evaluate our method on a two-channel 2D system, alanine dipeptide (ADP), and chignolin. Experimental and model-configuration details are provided in \cref{app:experimental-setup,app:2d,app:adp,app:chignolin-setup}.

\subsection{2D Two-Channel Potential}

\begin{figure*}[!t]
\centering
\includegraphics[width=\textwidth]{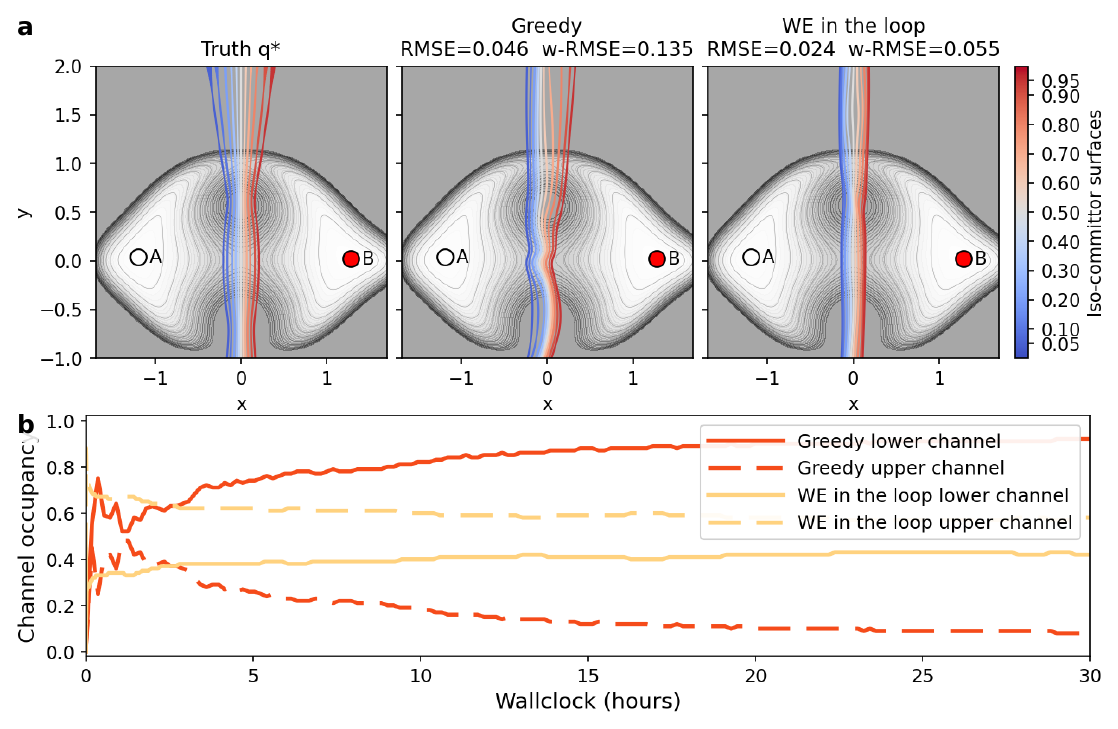}
\caption{\textbf{2D committor accuracy and channel occupancy.} (a) Finite-difference ground truth $q^{*}$ and best-checkpoint iso-committor surfaces for Q-greedy and WE-in-the-loop, annotated with RMSE and w-RMSE (RMSE weighted by the committor, giving more importance to the transition regions). (b) Lower- and upper-channel occupancy versus wall-clock time. Q-greedy progressively concentrates in the lower channel, whereas WE-in-the-loop maintains coverage of both.}
\label{fig:2d-accuracy-channels}
\end{figure*}

\paragraph{Why this system.} The exploration--exploitation tradeoff is hard to isolate in molecular systems, where the transition path ensemble is unknown and failures can have many causes. We therefore start with a 2D potential designed to make this tradeoff explicit. The potential, introduced by \citet{chen_efficient_2005} and studied by \citet{zhang_weighted_2009}, has two metastable states connected by two distinct reaction channels, with the asymmetry between the channels controlled by a parameter $\alpha$ (\cref{toy_potential_eqn}, \cref{fig:2d-accuracy-channels}a). A correct committor must account for both channels. A sampler that commits early to one channel will therefore learn a committor that is accurate along that channel but wrong along the other. \citet{zhang_weighted_2009} showed that for large $\alpha$, greedier transition path sampling methods become trapped in one channel while WE continues to sample both. This makes the system a natural test of whether the same failure carries over to committor learning, and whether WE-in-the-loop avoids it. Because the system is two-dimensional, the exact committor can also be computed by finite differences, so accuracy can be measured directly rather than through proxies.

\paragraph{Setup.} Following \citet{zhang_weighted_2009}, we set $\alpha = 72$ and define states $A$ and $B$ as open balls of radius $0.25$ centered at the two well minima. Both WE-in-the-loop and Q-greedy simulate overdamped Langevin dynamics (\cref{eq:overdamped}) at $\beta = 1$ with Euler--Maruyama step size $\Delta t = 10^{-3}$. The ground-truth committor $q^*$ is obtained by finite differences (\cref{app:2d}), and the full setup is given in \cref{app:2d-setup}.

\paragraph{Results.} At their best checkpoints, both methods learn a reasonable committor, but Q-greedy is less accurate in the transition region (\cref{fig:2d-accuracy-channels}a). The channel occupancy in \cref{fig:2d-accuracy-channels}b explains why: Q-greedy concentrates increasingly on the lower channel, while WE-in-the-loop continues to sample both. This gap widens as training continues. Because Q-greedy's data increasingly comes from a single channel, its sampling bias compounds, and at convergence it reaches an RMSE of $0.129$, compared with $0.032$ for WE-in-the-loop, decreasing significantly from its best value show in \ref{fig:2d-lowest-loss}, which is a snapshot when the model was performing best. Training loss alone does not reveal this failure. Early in training, before transition-region data has been collected, the loss can be trivially small for both methods (\cref{fig:2d-lowest-loss}
). These results show that principled exploration is necessary to maintain a correct committor estimate, exactly the failure the potential was designed to expose.

\subsection{Alanine Dipeptide}
\label{sec:exp:adp}

\begin{figure*}[t]
\centering
\includegraphics[width=\textwidth]{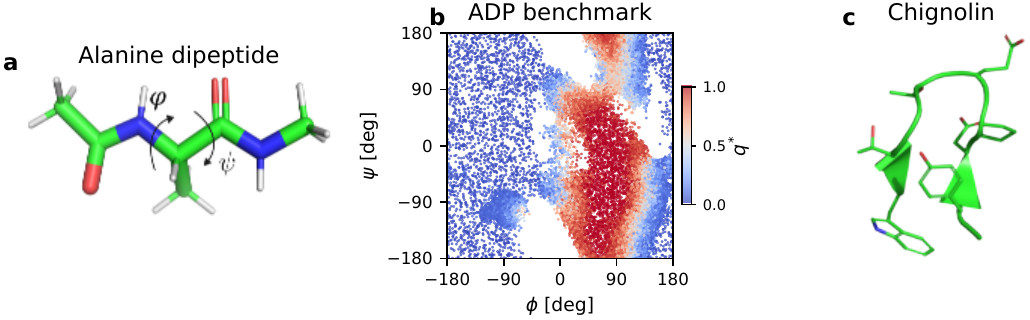}
\caption{\textbf{Molecular systems and ADP committor benchmark.} (a) Alanine dipeptide, with the backbone dihedrals $\phi$ and $\psi$ labeled. (b) Ramachandran plot of the ADP benchmark, with each conformation colored by its empirical committor $q^{*}$. (c) Chignolin structure used for the explicit-solvent simulations (solvent and ions omitted).}
\label{fig:system-overview}
\end{figure*}

\paragraph{Why ADP}
Alanine dipeptide (ADP) is a 22-atom peptide whose conformational dynamics are well described by its two backbone dihedral angles, $\phi$ and $\psi$ (\cref{fig:system-overview}a). Its transition between the $C_{7\mathrm{eq}}$ and $C_{7\mathrm{ax}}$ conformers is a standard benchmark for rare-event sampling and committor learning: the transition is slow enough to require enhanced sampling, it proceeds through more than one channel in $(\phi, \psi)$ space, and the system is small enough that reference committor values can be estimated by brute force. We define state $A$ as $C_{7\mathrm{eq}}$ and state $B$ as $C_{7\mathrm{ax}}$, using the dihedral ranges given in \cref{app:adp}.

\paragraph{Committor benchmark.}
\label{sec:exp:adp:dataset}
Evaluating committor learning methods on molecular systems is difficult because the true committor is unknown. To provide a common reference, we release a dataset of empirical committor values for approximately $360{,}000$ ADP conformations (\cref{fig:system-overview}b). Starting from approximately $10{,}000$ transition path sampling trajectories, we deduplicate configurations, stratify them by $(\phi, \psi)$, and launch $1{,}000$ independent Langevin trajectories from each selected conformation. The empirical committor is $q^*(x) = n_B(x) / (n_A(x) + n_B(x))$, where $n_A(x)$ and $n_B(x)$ count the trajectories that first reach $A$ and $B$, respectively. These labels are used only for evaluation and never for training. Full details of the pipeline are given in \cref{app:adp-dataset}.

\paragraph{WELP recovers missing channels where Q-greedy fails.}
\label{sec:exp:adp:results}
In standard ADP benchmarks, initial configurations are typically chosen close to all known transition channels, so the sampler never has to discover a pathway on its own. This assumption does not hold for larger systems, where the relevant pathways are not known in advance. To test exploration directly, we construct an adversarial setting in which initial configurations are biased away from one of the transition channels. We use overdamped dynamics here because, under underdamped dynamics, randomized initial velocities alone are enough to traverse the starting basin. In that case even a small set of starting positions effectively covers the full basin boundary (\cref{fig:adp-short-md}, \cref{app:adp}). The adversarial overdamped setting therefore isolates the question of interest: can a method recover a transition channel that is absent from its initial samples?

\begin{figure*}[t]
\centering
\includegraphics[width=\textwidth]{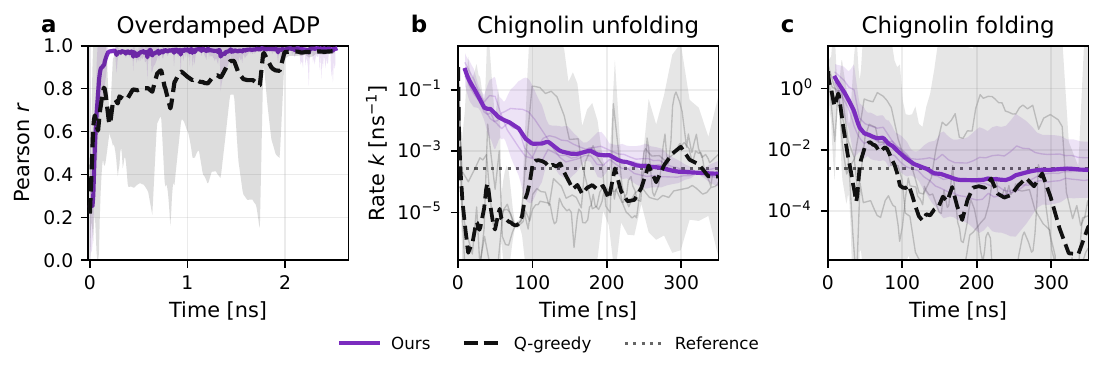}
\caption{\textbf{Overdamped ADP committor accuracy and chignolin rate convergence.} (a) Test-set Pearson $r$ versus aggregate simulation time for adversarial overdamped ADP. Both WE-in-the-loop and Q-greedy retrain at every iteration. Curves show means over $n=3$ seeds; bands show 95\% confidence intervals. (b,c) Chignolin unfolding and folding rate estimates from the Hill relation over 350~ns of aggregate simulation. Bold curves show geometric means over $n=3$ seeds, shaded bands show two-sided 95\% $t$-intervals in log-rate space, and thin curves show individual seeds. Dotted lines mark published reference rates. WELP always converges to the right rate at a significant speedup over Q-greedy.}
\label{fig:adp-convergence}
\label{fig:chignolin-rates}
\end{figure*}

WE-in-the-loop quickly reaches a high Pearson correlation with the benchmark committor, while Q-greedy converges much more slowly (\cref{fig:adp-convergence}a). \Cref{fig:adp-ovd-ramachandran} shows why. Q-greedy concentrates its sampling in one region of the Ramachandran plane and its committor estimate degrades elsewhere, whereas WE-in-the-loop continues to sample both channels. Notably, WE-in-the-loop performs comparably to the non-adversarial underdamped setting, which shows that it is robust to poor initialization. Explicit exploration therefore matters most when the initial samples do not already cover all relevant pathways, which is the expected regime for larger systems.

\paragraph{WELP is more efficient than WE alone}
We also evaluate our method in the standard underdamped setting to show the strength of our method in this setting. We run a set of ablations in this setting to quantify the speed up that our method has over using a human crafted CV in weighted ensemble. We find that WELP converges to the correct answer in less than 5 ns of simulation, while using WE only with RMSD only or even $q_\theta$ + RMSD requires up to 40 ns of simulation to converge (\cref{fig:adp-cv-ablations}). This suggests that our iterative approach has clear advantages over human crafted CVs; details are given in \cref{app:adp-ablations}. \cref{fig:adp-cv-ablations} shows that WELP converges faster than the greedy baselines in this setting as well. 

\subsection{WELP Converges Faster on the Higher-Dimensional Solvated Chignolin System}
\label{sec:exp:chignolin}

\paragraph{Setup.} To test whether WELP scales beyond small peptides, we study the folding and unfolding of chignolin, a 10-residue $\beta$-hairpin and one of the fastest-folding proteins known (\cref{fig:system-overview}c). Unlike ADP, its folding cannot be described by a small set of dihedral angles. The protein has 166 atoms, and we simulate it with 5,310 TIP3P water molecules and two sodium ions, giving 16,098 atoms in total. At 340~K, folding and unfolding occur on microsecond timescales \citep{lazzeri_molecular_2023}, well beyond what unbiased simulation can sample repeatedly at reasonable cost. We define the folded state A and the unfolded state B using Q, a smooth measure of the fraction of 74 native contacts, with A defined by $Q \geq 0.99$ and B by $Q \leq 0.01$, and simulate underdamped Langevin dynamics in OpenMM. Full details are given in \cref{app:chignolin-setup}.

Because no ground-truth committor is available for chignolin, we evaluate each method by the transition rates its learned committor implies. Let $\Phi_A$ be the flux of trajectories leaving $A$ through its boundary $\partial A$, and let $\langle \cdot \rangle_{\partial A}$ denote an average over the configurations at which they cross. A trajectory leaving $A$ reaches $B$ before returning to $A$ with probability $q_\theta$, so
\begin{equation}\label{eq:chignolin_rates}
    k_{AB} = \Phi_A \, \langle q_\theta \rangle_{\partial A},
    \qquad
    k_{BA} = \Phi_B \, \langle 1 - q_\theta \rangle_{\partial B}.
\end{equation}
We compare these estimates with published reference rates at 340~K \citep{lazzeri_molecular_2023}.

\paragraph{Results.} WELP's rate estimates converge to the reference values for both folding and unfolding within 350~ns of aggregate simulation (\cref{fig:chignolin-rates}b,c), less than the time of a single [folding/unfolding] event in unbiased simulation. Q-greedy's estimates remain far from the reference over the same simulation budget. Because the rates depend on $q_\theta$ only through its values at the basin boundaries, this agreement indicates that WELP learns an accurate committor where it matters most for kinetics: at the start of the transition, in a system with thousands of degrees of freedom.

\section{Discussion}

We introduced WE-in-the-loop (WELP), a framework that couples weighted ensemble sampling with neural network committor learning. The two components are mutually reinforcing. WE generates the short, statistically weighted trajectories that the semigroup loss requires, and the learned committor in turn provides an increasingly informative progress coordinate for WE. Our central finding is that exploration matters for committor learning, much as it does in reinforcement learning. Because a learned committor shapes its own training data, a sampler must balance exploiting the current estimate against exploring regions where that estimate may be wrong. WE achieves this balance in a principled way: adaptive binning along the committor maintains coverage across configuration space, bottleneck splitting concentrates computation in the transition region, and weight conservation keeps the resulting data statistically unbiased.

This balance is reflected across systems of increasing complexity. On the 2D two-channel potential, WELP maintains coverage of both transition channels throughout training and converges to an accurate committor. On alanine dipeptide with adversarial initialization, WELP recovers a transition channel absent from its initial samples, reaching a Pearson correlation of $0.934$ with the benchmark committor (vs.\ $0.732$ for the greedy baseline), and performing comparably to the standard, well-initialized setting. Even in that easier underdamped setting, WELP converges reliably with low variance across seeds. On chignolin in explicit solvent, WELP's folding and unfolding rate estimates converge to published reference values within 350~ns of aggregate simulation. Together, these results suggest that the benefits of principled exploration grow with system complexity, precisely the regime where initial samples are least likely to cover all relevant pathways.

Finally, our empirical ADP committor benchmark provides a common, training-independent reference for evaluating committor learning methods on a molecular system. We hope it enables more direct comparisons between methods than proxy metrics such as loss values or downstream rate estimates alone.

\section{Limitations and Future Work}

WELP bins walkers along a single learned coordinate, so distinct pathways with similar committor values share bins. Binning on additional coordinates, or on features learned jointly with the committor, could further improve coverage in systems with many competing channels. Each WE iteration also requires $K$ child rollouts per bin, and adapting $K$ to the local uncertainty of the committor could reduce this cost. More broadly, the closed loop between sampling and learning is not specific to the committor: any learned reactive quantity that can define bins or interfaces could guide sampling in the same way, which is a promising direction for scaling to larger biomolecular systems. Our evidence spans exact 2D ground truth, ADP under two dynamical regimes, and atomistic chignolin, but larger proteins and complex systems remain for future work. 
\section{Data and Code Availability}
\label{sec:data-code}

The WELP implementation and example simulation inputs are publicly
available at \url{https://github.com/jackychen08/WELP}.


\section*{Impact Statement}

This paper presents work whose goal is to advance the field of Machine
Learning. There are many potential societal consequences of our work, none
which we feel must be specifically highlighted here.

\FloatBarrier

\bibliographystyle{unsrtnat}
\bibliography{references}


\appendix

\section{Semigroup Loss Derivation}
\label[appendix]{app:losses}

\subsection{Semigroup loss}
\label[appendix]{app:loss-semigroup}

Solving for the committor directly as a regression problem would require
committor labels in the transition region, which we do not have. Following
\citet{mitchell_committor_2024} we employ a semigroup approach that
instead uses a Markov self-consistency identity for $q$ that can be
estimated from short rollouts of the dynamics.

Let $\boldsymbol{X}_t^{\boldsymbol{x}}$ denote the dynamics initialized at $\boldsymbol{x}$ and
$\sigma = \inf\{t\geq 0 : \boldsymbol{X}_t \in A \cup B\}$, i.e.\ the first basin hit.
By the Markov property, for any $\tau > 0$,
\begin{equation}
    q(\boldsymbol{x})
    = \mathbb{E}\!\left[q\bigl(\boldsymbol{X}_{\tau \wedge \sigma}^{\boldsymbol{x}}\bigr)\right],
    \quad \boldsymbol{x} \in \Omega \setminus (A \cup B).
\end{equation}
The stopping at $\sigma$ is what enforces the boundary conditions
$q|_{\partial A}=0$, $q|_{\partial B}=1$ into the identity.
Trajectories that have hit $B$ first contribute $1$, while those that have
hit $A$ first contribute $0$, and those still in the transition region
contribute $q$ at their current configuration. The stopped operator
$\tau \wedge \sigma$ indicates that we either simulate for $\tau$ time
or until we hit a boundary condition $\sigma$. Intuitively, if a
configuration is evolved forward in many separate runs and one looks at
where each run lands, the committor of the original configuration is
the average committor over those resulting ``child'' configurations.

\citet{mitchell_committor_2024} and \citet{li_semigroup_2022} turn this semigroup
identity into a self-consistent training objective. From each
$x$, $N$ short trajectories are run with adaptive termination
on basin entry, and is approximated by
\begin{equation*}
    y_\tau(x) \;=\; \frac{1}{N}\sum_{i=1}^{N} q_\theta\bigl(\boldsymbol{X}^{(i)}_{\tau \wedge \sigma_i}\bigr),
\end{equation*}
where $q_\theta$ is replaced by $0$ or $1$ on trajectories terminated
in $A$ or $B$. We use the symmetrized log form of the residual, with a
stop-gradient $\mathrm{sg}[\cdot]$ on the targets for training stability,
giving the semigroup loss $\mathcal{L}_{\mathrm{sm}}$ in
\cref{eq:semigroup_loss}. The symmetrization treats the committor in the forward direction
 $q$ and the committor in the reverse direction $1-q$ on equal footing.

\section{Experimental Setup}
\label[appendix]{app:experimental-setup}

In this section we evaluate our WE in the loop method on a two-channel 2D system, alanine dipeptide (ADP), and Chignolin. We show that a greedy approach becomes biased while our method does not. We also curate and release (upon publication) a dataset of empirically determined committor values for underdamped Langevin dynamics on ADP. Full model architectures for both the greedy baseline and the committor used by our WE in the loop method on each system are in App.~\ref{app:2d} (2D) and App.~\ref{app:adp} (ADP). On ADP, both methods use the same pairwise-distance input, while our committor network adds residual connections.

\footnotesize $^{*}$The q-logit + WSM variant uses three seeds; other lower-training-ratio WE variants update every 10 combined iterations. The 2D panel reports selected checkpoints. Within each setting, methods share dynamics, inputs, evaluation data, and simulation-time accounting.

\subsection{Shared Training Implementation}
\label[appendix]{app:training-implementation}

To supply the
$\mathcal{L}_{\mathrm{sm}}$ targets, we store parent--children pairs in a first-in-first-out buffer. Within each training call, a target network fixes child values and is then updated from the current network. Cadence and optimizer settings are experiment-specific; architectures follow \citet{mitchell_committor_2024}, with residual connections added between layers.

\section{Two-Channel 2D System}
\label[appendix]{app:2d}

\subsection{Setup and ground truth}
\label[appendix]{app:2d-setup}

\textbf{Geometry.} The two-channel ZJZ potential of
\cref{toy_potential_eqn},
\begin{align}\label{toy_potential_eqn}
\frac{U(x, y)}{k_B T} &= \alpha (x^2 + y^2 - 1)^2 y^2 \\ \nonumber
  &\quad - \exp\{-4[(x-1)^2 + y^2]\} \\ \nonumber
  &\quad - \exp\{-4[(x+1)^2 + y^2]\} \\ \nonumber
  &\quad + \exp[8(x - 1.5)] + 4\exp[-8(x + 1.5)] \\ \nonumber
  &\quad + \exp[-4(y + 0.25)] + 16\exp(-2x^2),
\end{align}
has two metastable basins near $(\pm 1, 0)$
connected by a \emph{direct} channel through $(0,0)$ at $y\!\approx\!0$
and an \emph{upper} channel arcing over $y\!\approx\!1$, separated by an
inter-channel ridge at $(0,0.5)$. Well minima are located numerically
once at module-load by 1000-step SGD with momentum (lr $10^{-2}$, $\beta
0.9$) and pinned to $W_A=(-1.196533, 0.037285)$ and
$W_B=(1.279177, 0.021035)$. Basins $A,B$
are defined as balls of radius $r_{\mathrm{basin}}=0.25$ around these well
minima.

\textbf{Reference SDE settings.} All experiments integrate
\cref{overdamped_eq} via the Euler--Maruyama update below with
$\beta=\gamma=1$ and $\Delta t=10^{-3}$. The small $\Delta t$ is required
for numerical stability against the steep $e^{8(x-1.5)}$ wall.
We use overdamped dynamics to keep the committor a
function of $(x,y)$ only, admitting a grid solution for $q^{*}$.

\textbf{Euler--Maruyama update.} With $\gamma=1$, the integrator update is
\begin{equation}
x_{t+\Delta t} = x_t - \nabla U(x_t)\Delta t
+ \sqrt{2\beta^{-1}\Delta t}\xi_t,\;\xi_t\sim\mathcal{N}(0,I_2).
\label{eq:em-2d}
\end{equation}

\textbf{Ground-truth Boundary Value Problem (BVP).} On the uniform $401\!\times\!401$ grid we
discretize the advection--diffusion BVP $\mathcal{L}_{\mathrm{OD}} q = 0$
with $\mathcal{L}_{\mathrm{OD}} = -\nabla U \cdot \nabla + \beta^{-1}\Delta$ and Dirichlet boundary conditions $q|_A=0,\,q|_B=1$: centered
differences for advection, the standard 5-point Laplacian for diffusion,
and one-sided ghost-point mirroring at the box edge (which collapses the
centered-advection term and turns the Laplacian into
$2(q_1-q_0)/\Delta x^{2}$, i.e.\ a zero-flux Neumann condition).
Dirichlet rows ($q\!=\!0$ in $A$, $q\!=\!1$ in $B$) replace the operator
on basin pixels. The resulting sparse $N\!\times\!N$ system
($N\!=\!160{,}801$) is solved once with
\texttt{scipy.sparse.linalg.spsolve}; analytic $\nabla U$ is hard-coded
for speed and conditioning. The output $q^{*}$ is clipped to $[0,1]$.

\textbf{Channel labeling and occupancies.} A point $(x,y)$ is labeled
\emph{upper} if $|x|<0.7$ and $y>0.5$, \emph{lower} if $|x|<0.7$ and
$y\le 0.5$. Reported $f_{\mathrm{upper}}, f_{\mathrm{lower}}$ are
fractions over labeled points only.

\subsection{Greedy seed-based baseline}
\label[appendix]{app:2d-grant}

\textbf{Network and training schedule.} The committor net for the 2D system is a 2-layer
MLP $2\!\to\!50\!\to\!1$ with $q=\sigma(\mathrm{net})$. We run
$K\!=\!3000$ outer iterations with $E\!=\!100$ inner Adam epochs per
iteration.

\textbf{Greedy seed selection.} On the very first iteration, or whenever the
previous swarm successfully transited, the seed is drawn with replacement from
the cached boundary-flux exit configurations.

\textbf{Loss gradient stability.} The log-loss is evaluated only on
points whose $q^{\mathrm{t}}\!\in\!(0,1)$, with predictions clamped to
$[\varepsilon,1-\varepsilon]$, $\varepsilon\!=\!10^{-6}$, before the log
to keep gradients finite when $\sigma$ saturates. Inner optimization is
Adam at lr $10^{-2}$, mini-batch size $64$ over the seed buffer, with
gradient-norm clipping at $10.0$.

\subsection{WE in the loop method}
\label[appendix]{app:2d-we}

\textbf{Simulations and progress coordinate.} Two WESTPA~2.0
simulations run concurrently with target-state recycling, with one from both directions. 
Each walker records
$(q,\mathrm{rmsd}_B,\mathrm{in\_target})$, with $q$ filled in by the
network at each WE step. The committor, $q$, is used as the collective variable. Minimum adaptive binning (MAB) \cite{torrillo_minimal_2021}is applied on the collective variable to encourage more simulation at bottleneck nodes. 

\textbf{Network.} The committor net is identical to the
greedy baseline ($2\!\to\!50\!\to\!1$) with $q=\sigma(\mathrm{net})$.

\textbf{Basis states.} 10 jittered points per basin (Gaussian
$\sigma=0.03$ around the well minimum, rejection-sampled to lie inside
$r_{\mathrm{basin}}$), each given uniform initial weight $1/n$.

\begin{figure}[!htbp]
\centering
\includegraphics[width=\linewidth]{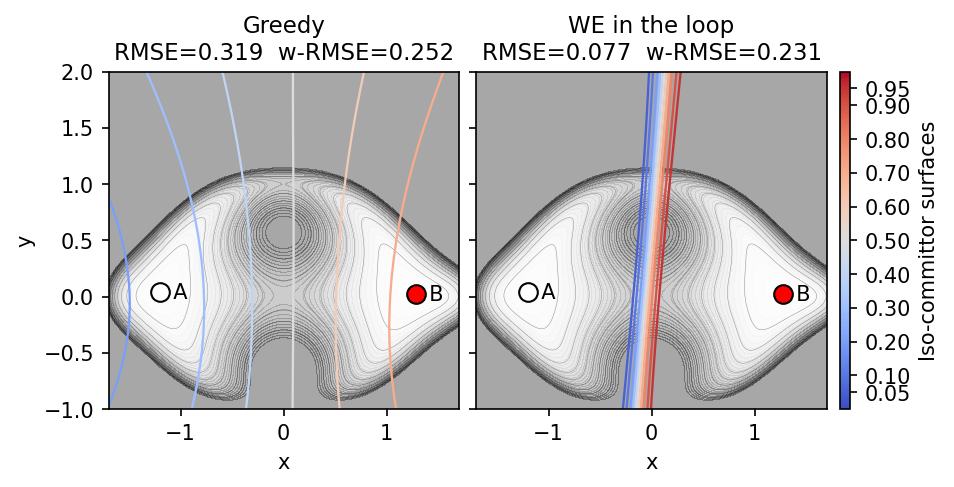}
\caption{\textbf{Iso-committor surfaces at the lowest-loss checkpoint (2D).} Greedy (left) and WE in the loop (right) at the iteration with the smallest training loss. Because data is still being collected at this point, both methods can trivially satisfy the loss in regions that have not yet been sampled; the lowest-loss checkpoint therefore does not correspond to the most accurate committor.}
\label{fig:2d-lowest-loss}
\end{figure}

\section{Alanine Dipeptide System}
\label[appendix]{app:adp}

This appendix groups everything ADP-specific: how the benchmark
dataset was built (\cref{app:adp-dataset}) and how the committor in our WE in the loop
method was trained (\cref{app:adp-we}).

\subsection{Dataset generation}
\label[appendix]{app:adp-dataset}

\textbf{Transition path sampling.}
\label[appendix]{app:adp-tps}
Transition path sampling for alanine dipeptide was conducted via the
OpenPathSampling library, interfaced with the OpenMM simulation engine
\cite{eastman_openmm_2024}. Molecular dynamics trajectories were
propagated using a Langevin integrator employing the middle
discretization scheme, with a time step of $\Delta t = 1~\text{fs}$ and
a friction coefficient of $\gamma = 1~\text{ps}^{-1}$. All simulations
employed the CHARMM27 force field with hydrogen bond length constraints
enforced throughout. Conformational snapshots were recorded at
intervals of 10 fs. Metastable basins were defined in the
$(\phi, \psi)$ dihedral angle space following the convention
established in \cite{mitchell_committor_2024}. Specifically, basin
$\mathcal{A}$ is centered at $[\phi, \psi] = [-150^\circ, 170^\circ]$
and basin $\mathcal{B}$ at $[\phi, \psi] = [90^\circ, -50^\circ]$,
with each region defined as a square of side length $20^\circ$ in the
Ramachandran plane.

\textbf{Source trajectories and deduplication.}
We start from $\sim 10{,}000$ TPS trajectories of alanine dipeptide in
vacuum (CHARMM27). Frames are concatenated and
deduplicated to unique conformers by rounding atom positions to
$10^{-3}$~nm precision before pasing through \texttt{torch.unique}.

\textbf{Stratified subsampling.}
Conformers are placed into a $5^{\circ}\!\times\!5^{\circ}$ grid on the
Ramachandran plane ($72{\times}72$ bins). Within each occupied bin: if
the bin contains $\le 100$ conformers, all are kept; otherwise, the
all-atom pairwise-distance matrix is fed to agglomerative clustering
with complete linkage and \texttt{n\_clusters}$=100$, and from each
cluster we keep the conformer nearest to the cluster mean. For bins
exceeding $10{,}000$ conformers we first reduce to $10{,}000$ centers
with MiniBatchKMeans (batch size 4096, 3 inits) before agglomerative
clustering. Since complete-linkage scales quadratically and is the
runtime/memory bottleneck, we employ K-Means as pre-filter. 

\textbf{State definitions.}
The two metastable states are centered at
$A=(\phi,\psi)=(-150^{\circ},170^{\circ})$ and
$B=(90^{\circ},-50^{\circ})$, both with a $\pm 10^{\circ}$ acceptance
window in each angle.

\textbf{Brute-force MD.}
For each selected representative we propagate $N_{\mathrm{trials}}=1000$
independent Langevin trajectories using OpenMM's
\texttt{LangevinMiddleIntegrator} ($T=300$~K, friction
$\gamma=0.3$~ps$^{-1}$, timestep $\Delta t=1$~fs. Initial velocities for each trial
are independently drawn from the Maxwell--Boltzmann distribution at
$300$~K. Every $100$ steps ($0.1$~ps) we recompute $(\phi,\psi)$ and
check membership in $A$ or $B$; the first hit terminates the trial.
Trajectories that fail to commit are extended in $10^{6}$-step batches
up to a $10^{8}$-step ($100$~ns) safety cap and recorded as unresolved.

\textbf{Empirical committor.}
For each conformer $x$, the empirical committor is
\begin{equation}
q^{*}(x) \;=\; \frac{n_B(x)}{n_A(x) + n_B(x)},
\label{eq:adp-empirical-committor}
\end{equation}
where $n_A,n_B$ are counts of trials terminating in $A$ and $B$
respectively. Unresolved trials are excluded from the denominator. 

\textbf{Balanced split.}
The phi/psi-stratified pipeline above produces a dataset whose
\emph{committor} distribution is heavily skewed toward $q^{*}\!\approx\!0$
and $q^{*}\!\approx\!1$, since the basins occupy most of the populated
Ramachandran area. To support evaluation across the full committor
range, we additionally subsample to obtain a balanced subset: starting from the
conformers with valid empirical committor, we partition $[0,1]$ into
$20$ bins of width $0.05$, take the count of the smallest populated
bin as the target per bin, and uniformly downsample each populated bin
to that count (seed $0$, sampling without replacement). The resulting
balanced split has equal mass across all populated $q^{*}$ bins,
including the transition region.

\subsection{WE in the loop method}
\label[appendix]{app:adp-we}

\textbf{System, states, and directional simulations.}
The ADP experiments use the 22-atom peptide in vacuum with a CHARMM27 GROMACS topology, no nonbonded cutoff, and constrained hydrogen bonds. The states are the $20^{\circ}\!\times20^{\circ}$ squares centered at $A=(-150^{\circ},170^{\circ})$ and $B=(90^{\circ},-50^{\circ})$. Separate $A\!\to\!B$ and $B\!\to\!A$ WESTPA simulations recycle walkers at the target state. Non-target walkers are binned by MAB on $q$ alone with 20 walkers per occupied bin, followed by splitting and merging the weights. 

\textbf{Overdamped protocol.}
The adversarial experiment uses OpenMM's Brownian integrator at 300~K with friction $1~\mathrm{ps}^{-1}$ and a $0.01$-fs step. Each WE segment contains 1,000 steps (10~fs) and stores 11 frames at 1-fs intervals. Each directional simulation starts from 20 equally weighted boundary configurations from the restricted initial ensemble by removing starting points from half the initial perimeter as shown in \cref{fig:adp-short-md}. The network is updated after each newly completed WE iteration, and each directional simulation is run for at most 460 iterations.

\textbf{Underdamped protocol.}
The frequently trained WE experiment uses OpenMM's Langevin-middle integrator at 310~K with friction $0.3~\mathrm{ps}^{-1}$ and a 1-fs step. Each WE segment contains 100 steps (100~fs), stores 11 frames at 10-fs intervals, and redraws velocities from the 310-K Maxwell--Boltzmann distribution at the start of every segment. Each directional simulation starts from 10 equally weighted boundary configurations. The network is updated after each newly completed WE iteration, and each directional simulation is run for at most 4,500 iterations.

\textbf{Network.} The committor net is the residual MLP variant of
the \citet{mitchell_committor_2024} pairwise-distance architecture:
$45$ pairwise interatomic distances on $10$ heavy atoms feed a
$4$-block residual trunk with embed dim $64$ and FFN hidden $128$
($3$ residual layers per block), followed by a sigmoid head
$q=\sigma(\mathrm{net})$. The greedy baseline uses the same
input and model architecture, except for the residual connections in our committor net.

\textbf{Optimizer and replay.} We utilized Adam at lr $10^{-3}$, batch size
$1024$, $E\!=\!120$ inner epochs per training call, with the a FIFO
replay buffer.

\textbf{WE configuration.} Two WESTPA simulations are run with
target-state recycling. Adaptive binning uses MAB on $q$ alone inside an
outer recycling shell, matching the
2D pipeline in \cref{app:2d-we}.

\textbf{Within-bin merge and re-split (WSM).} We force walkers within each bin to merge before WESTPA restores its target count. The resulting siblings per parent reduce Monte Carlo noise in the semigroup target.

\subsection{Why overdamped dynamics for the adversarial setting}
\label[appendix]{app:adp-short-md}

\begin{figure}[t]
\centering
\begin{subfigure}{0.48\columnwidth}
  \centering
  \includegraphics[width=\linewidth]{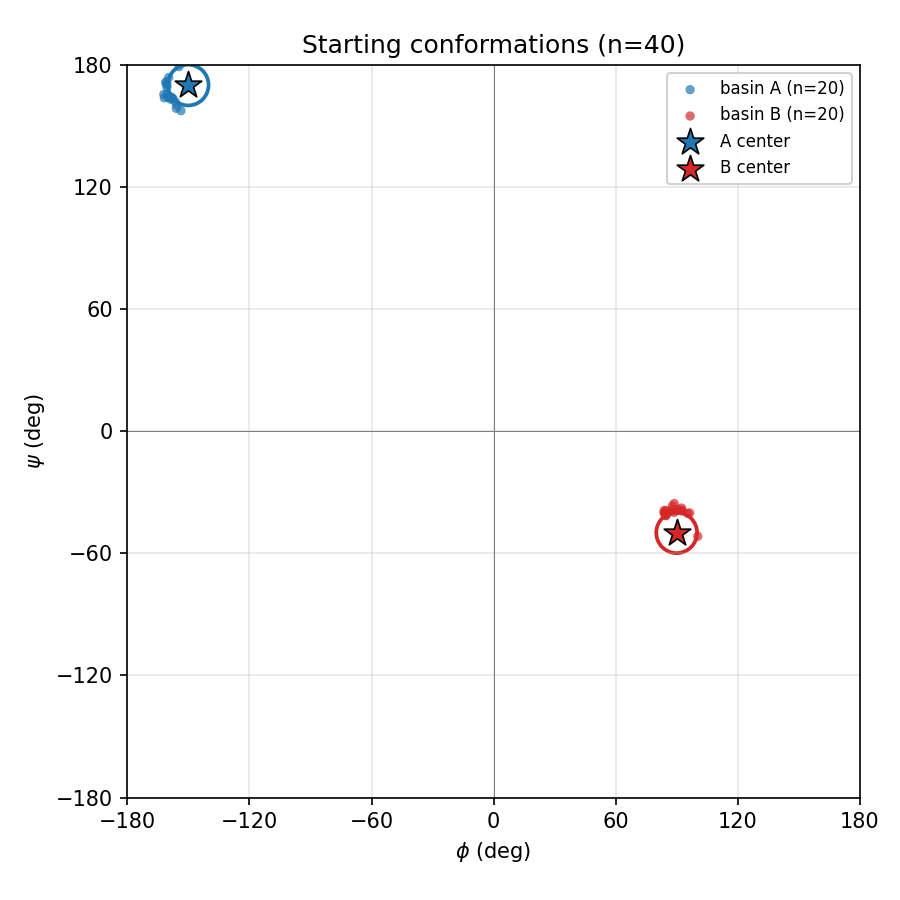}
  \caption{Starting conformations ($n\!=\!40$).}
  \label{fig:adp-short-md-start}
\end{subfigure}
\hfill
\begin{subfigure}{0.48\columnwidth}
  \centering
  \includegraphics[width=\linewidth]{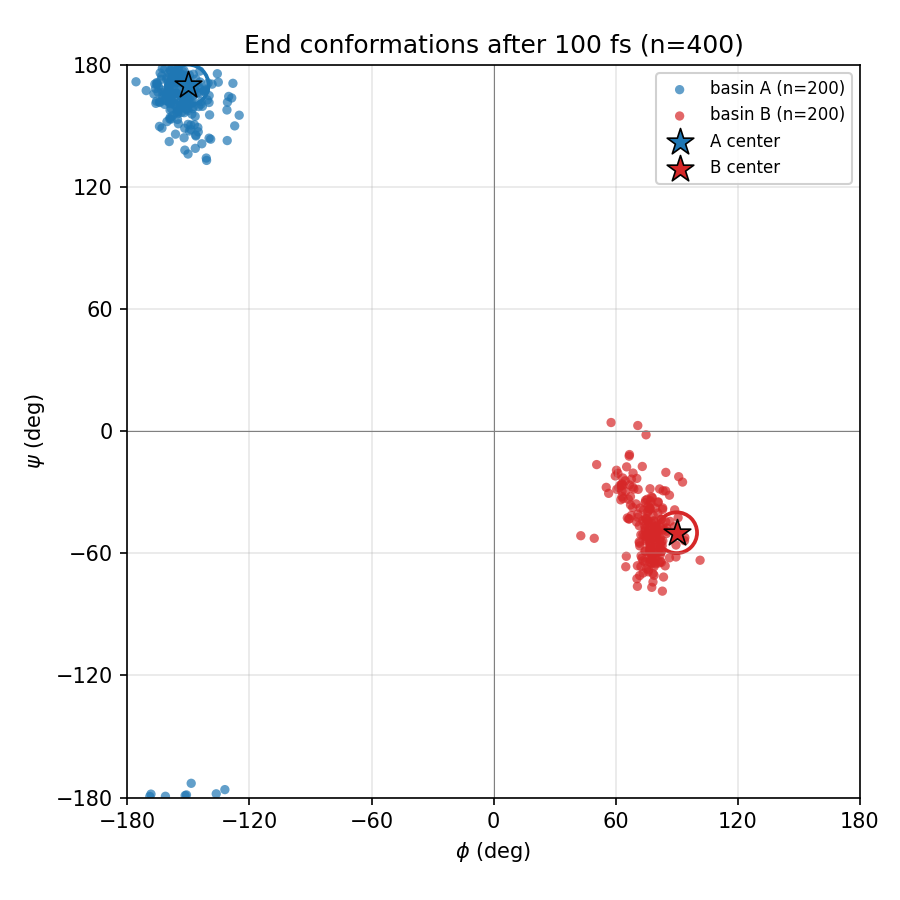}
  \caption{Endpoints after $100$~fs underdamped MD ($n\!=\!400$).}
  \label{fig:adp-short-md-end}
\end{subfigure}
\caption{\textbf{Underdamped Langevin dynamics rapidly spread starting
conformations across the source basin.} (a) Forty starting conformations
sampled tightly around the basin-$A$ ($C7_{\mathrm{eq}}$, blue) and
basin-$B$ ($C7_{\mathrm{ax}}$, red) centers. (b) Endpoint $(\phi,\psi)$
of $400$ trajectories after a single $100$~fs underdamped propagation
with freshly randomized initial velocities. The endpoint spread already
covers the basin extent, so spatial biasing of initial points is
largely undone by velocity randomization plus a short propagation.
This is why the adversarial setting in \cref{sec:exp:adp:results} is
run with overdamped dynamics instead.}
\label{fig:adp-short-md}
\end{figure}

The adversarial overdamped setting in \cref{sec:exp:adp:results} restricts
the starting conformations to one half of the Ramachandran plane in order
to suppress one transition channel as shown in \cref{fig:adp-short-md}. We aim to bias the initial conditions to see whether or not sufficient exploration is built into the algorithm. This restriction is only meaningful
if the dynamics actually keep walkers near the prescribed initial
$(\phi,\psi)$ on the timescale of a single WE step. Under underdamped
Langevin dynamics the momenta are freshly resampled from the
Maxwell--Boltzmann distribution. Even a single WE-iteration's worth of propagation is equivalent to having started from the entire basin boundary; inertia carries walkers across the full basin extent within one short propagation, washing out any spatial biasing of the initial points. \cref{fig:adp-short-md} illustrates this
empirically. Forty conformations sampled tightly around the basin
centers spread to cover the full basin extent after a single 100~fs
underdamped step with randomized initial velocities. Therefore, switching to
overdamped dynamics removes this effect because it has no momentum
and lets the spatial restriction on initial conditions actually take
hold.

\begin{figure*}[t]
\centering
\includegraphics[width=\textwidth]{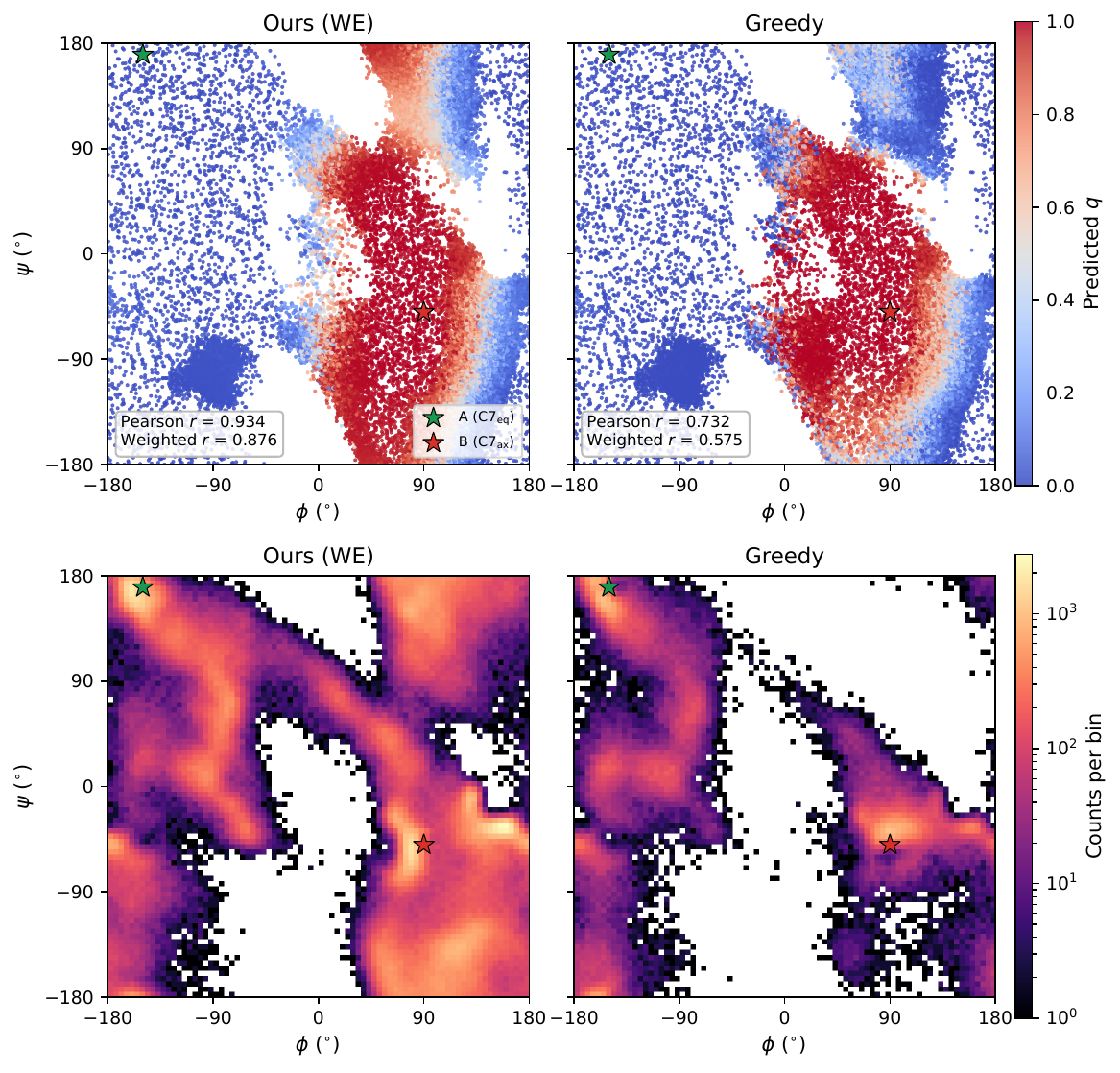}
\caption{\textbf{Overdamped ADP committor accuracy and sampling in the adversarial setting.} Top: predicted committor on the Ramachandran plane for WE in the loop and Greedy. Bottom: corresponding sampling density. Greedy collapses toward one transition channel ($r=0.732$, weighted $r=0.575$), whereas WE retains coverage of both routes ($r=0.934$, weighted $r=0.876$).} 
\label{fig:adp-ovd-ramachandran}
\end{figure*}

\subsection{Underdamped ADP comparisons}
\label[appendix]{app:adp-ablations}

\begin{figure}[H]
\centering
\includegraphics[width=0.97\textwidth]{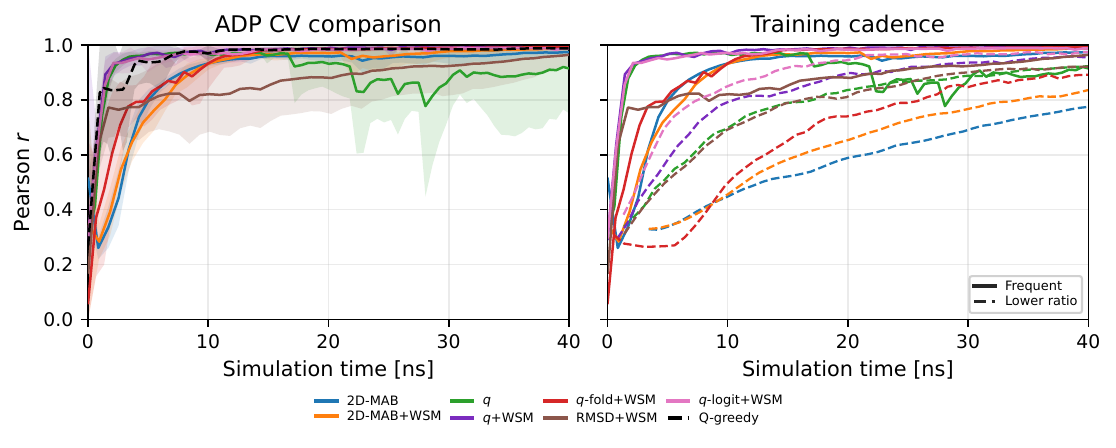}
\caption{\textbf{Underdamped ADP comparisons.} Test-set Pearson $r$ versus aggregate simulation time. (a) Collective-variable and WSM variants with frequent training; Q-greedy is the baseline. (b) Each WE method with frequent training (solid) and its lower-training-ratio counterpart (dashed). Curves are seed means; bands in (a) are 95\% confidence intervals.}
\label{fig:adp-cv-ablations}
\end{figure}

Panel (a) of \cref{fig:adp-cv-ablations} compares seven WE configurations. 2D-MAB bins jointly on the predicted $q$ and root mean square distance (RMSD) while 2D-MAB+WSM adds within-bin merge and re-split (WSM). The $q$ configuration bins on $q$ alone; $q$+WSM adds WSM and is our default. The $q$-fold+WSM configuration bins on $q(1-q)$, RMSD+WSM bins only on target-state RMSD, and $q$-logit+WSM bins on $\log[q/(1-q)]$. Q-greedy launches trajectories from the endpoint with the largest predicted commitment instead of resampling a weighted ensemble. WSM is used since it forces the splitting of walkers, which decreases the variance seen under the semigroup objective \ref{eq:semigroup_loss}. 

Panel (b) compares each WE configuration under two training schedules. The training-to-simulation ratio is the number of network-training calls per newly completed WE iteration, summed over the $A\!\to\!B$ and $B\!\to\!A$ simulations. Frequent training uses a nominal $1{:}1$ ratio, whereas the lower-ratio schedule uses $1{:}10$. We show that more frequent neural network training reduces the amount of molecular simulation required to reach convergence. This property becomes increasingly advantageous as we scale to larger systems, since model training is expected to scale more favorably with system size than molecular simulation. For smaller systems where training is more expensive than simulation, one can choose a less frequent training cadence. 

\section{Chignolin Setup}
\label[appendix]{app:chignolin-setup}

\paragraph{System and dynamics.}
We use model 1 of CLN025 (YYDPETGTWY; PDB 2RVD), add missing atoms and hydrogens at pH~7 with PDBFixer, and parametrize the 166-atom peptide with Amber ff99SB. The peptide is neutralized by two Na$^+$ ions and solvated with 5,310 TIP3P waters in a 5.5-nm cube, giving 16,098 atoms. We use PME with a 0.95-nm cutoff, constrained hydrogen bonds, and rigid water. After energy minimization and 1~ns of NPT equilibration at 340~K and 1~bar, the barostat is removed and the 5.528-nm box is fixed for NVT production. OpenMM's Langevin-middle integrator uses friction $1~\mathrm{ps}^{-1}$ and a 2-fs step.

\paragraph{System definition}
We define the folded and unfolded states using the smooth fraction
of native contacts,
\begin{equation}
Q(x) =
\frac{1}{74}
\sum_{k=1}^{74}
\frac{1}{
1+\exp\left[\beta\left(r_k(x)-\lambda r_k^0\right)\right]
}.
\end{equation}
Here, $r_k(x)$ is the periodic distance for native-contact pair $k$,
$r_k^0$ is its reference distance,
$\beta=50\,\mathrm{nm}^{-1}$, and $\lambda=1.8$.
The contact pairs and reference distances follow the archived
benchmark definition. Basin $A$ is folded ($Q\geq 0.99$), and
basin $B$ is unfolded ($Q\leq 0.01$). The structural coordinate
$Q$ is distinct from the learned committor $q_{\theta}$.

\paragraph{WE protocol.}
Separate steady-state $A\!\to\!B$ and $B\!\to\!A$ WESTPA simulations recycle at the target. Each segment comprises 1,000 steps (2~ps), stores 11 frames at 0.2-ps intervals, and redraws velocities at 340~K. The progress coordinate is again the learnt committot $q$, with adapative binning to encourage more simulation at key bottle necks. Six additional bins cover $q_\theta<0.01$ and $q_\theta>0.99$, to make sure simulations still cover boundary conditions. In the reactive region, we place ten logit-MAB bins between them and 20 walkers per bin. Without the additional bins, we observed that WELP still learns a committor that qualatively looks the same, but is less accurate in regions near A and B. The additional bins were placed to improve accuarcy near the basin, where the committor was used to evaluate the rate. 

\paragraph{Network and training.}
The input is all 13,695 pair distances among the 166 peptide atoms, standardized once using both fixed boundary ensembles. A 100-unit input layer feeds three 100-unit residual layers with LeakyReLU slope 0.01 and a sigmoid output. After 10 combined warm-up iterations, we train every from a 65,000-entry FIFO buffer. Each call uses the semigroup objective, a fixed target-network snapshot, and 120 AdamW epochs (learning rate $10^{-4}$, weight decay $10^{-4}$, batch size 1,024, gradient-norm cap 1).



\end{document}